\documentclass[11pt,a4paper]{article}
\usepackage[hyperref]{acl2020}
\usepackage{times}
\usepackage{latexsym}

\usepackage{microtype}
\usepackage{amsmath}
\usepackage{amsfonts}
\usepackage{graphicx}
\usepackage{graphicx}
\usepackage{comment}
\usepackage{caption}
\usepackage{subcaption}
\usepackage{booktabs}
\usepackage{threeparttable}

\usepackage{comment}

\aclfinalcopy 
\def\aclpaperid{2487} 

\title{Encoder-Decoder Models Can Benefit from Pre-trained \\ Masked Language Models in Grammatical Error Correction}

\author{Masahiro Kaneko$^{1,}$$^2$\quad Masato Mita$^{2,}$$^3$\quad Shun Kiyono$^{2,}$$^3$\quad Jun Suzuki$^{3,}$$^2$\quad Kentaro Inui$^{3,}$$^2$ \\
$^1$Tokyo Metropolitan University\\
$^2$RIKEN Center for Advanced Intelligence Project\\
$^3$Tohoku University\\
{\tt kaneko-masahiro@ed.tmu.ac.jp} \\
{\tt \{masato.mita, shun.kiyono\}@riken.jp} \\
{\tt \{jun.suzuki, inui\}@ecei.tohoku.ac.jp}}

\date{}

\begin{document}
\maketitle
\begin{abstract}

This paper investigates how to effectively incorporate a pre-trained masked language model (MLM), such as BERT, into an encoder-decoder (EncDec) model for grammatical error correction (GEC).
The answer to this question is not as straightforward as one might expect because the previous common methods for incorporating a MLM into an EncDec model have potential drawbacks when applied to GEC. 
For example, the distribution of the inputs to a GEC model can be considerably different (erroneous, clumsy, etc.) from that of the corpora used for pre-training MLMs; however, this issue is not addressed in the previous methods. 
Our experiments show that our proposed method, where we first fine-tune a MLM with a given GEC corpus and then use the output of the fine-tuned MLM as additional features in the GEC model, maximizes the benefit of the MLM.
The best-performing model achieves state-of-the-art performances on the BEA-2019 and CoNLL-2014 benchmarks.
Our code is publicly available at: \url{https://github.com/kanekomasahiro/bert-gec}.

\end{abstract}

\section{Introduction}

Grammatical Error Correction (GEC) is a sequence-to-sequence task where a model corrects an ungrammatical sentence to a grammatical sentence.
Numerous studies on GEC have successfully used encoder-decoder (EncDec) based models, and in fact, most current state-of-the-art neural GEC models employ this architecture \cite{zhao-etal-2019-improving, grundkiewicz-etal-2019-neural, kiyono-etal-2019-empirical}.

In light of this trend, one natural, intriguing question is whether neural EndDec GEC models can benefit from the recent advances of masked language models (MLMs) since MLMs such as BERT~\cite{devlin-etal-2019-bert} have been shown to yield substantial improvements in a variety of NLP tasks~\cite{Qiu2020PretrainedMF}.
BERT, for example, builds on the Transformer architecture \cite{NIPS2017_7181} and is trained on large raw corpora to learn general representations of linguistic components (e.g., words and sentences) in context, which have been shown useful for various tasks. 
In recent years, MLMs have been used not only for classification and sequence labeling tasks but also for language generation, where combining MLMs with EncDec models of a downstream task makes a noticeable improvement~\cite{lample2019cross}.

Common methods of incorporating a MLM to an EncDec model are initialization (init) and fusion (fuse).
In the init method, the downstream task model is initialized with the parameters of a pre-trained MLM and then is trained over a task-specific training set~\cite{lample2019cross,rothe2019leveraging}.
This approach, however, does not work well for tasks like sequence-to-sequence language generation tasks because such tasks tend to require a huge amount of task-specific training data and fine-tuning a MLM with such a large dataset tends to destruct its pre-trained representations leading to catastrophic forgetting~\cite{Zhu2020IncorporatingBI,McCloskey1989CatastrophicII}.
In the fuse method, pre-trained representations of a MLM are used as additional features during the training of a task-specific model~\cite{Zhu2020IncorporatingBI}.
When applying this method for GEC, what the MLM has learned in pre-training will be preserved; however, the MLM will not be adapted to either the GEC task or the task-specific distribution of inputs (i.e., erroneous sentences in a learner corpus), which may hinder the GEC model from effectively exploiting the potential of the MLM.
Given these drawbacks in the two common methods, it is not as straightforward to gain the advantages of MLMs in GEC as one might expect.
This background motivates us to investigate how a MLM should be incorporated into an EncDec GEC model to maximize its benefit. 
To the best of our knowledge, no research has addressed this research question. 

In our investigation, we employ BERT, which is a widely used MLM~\cite{Qiu2020PretrainedMF}, and evaluate the following three methods: (a) initialize an EncDec GEC model using pre-trained BERT as in \citet{lample2019cross} (BERT-init), (b) pass the output of pre-trained BERT into the EncDec GEC model as additional features (BERT-fuse)~\cite{Zhu2020IncorporatingBI}, and (c) combine the best parts of (a) and (b). 

In this new method (c), we first fine-tune BERT with the GEC corpus and then use the output of the fine-tuned BERT model as additional features in the GEC model.
To implement this, we further consider two options:
(c1) additionally train pre-trained BERT with GEC corpora (BERT-fuse mask), and 
(c2) fine-tune pre-trained BERT by way of the grammatical error detection (GED) task  (BERT-fuse GED).
In (c2), we expect that the GEC model will be trained so that it can leverage both the representations learned from large general corpora (pre-trained BERT) and the task-specific information useful for GEC induced from the GEC training data.

Our experiments show that using the output of the fine-tuned BERT model as additional features in the GEC model (method (c)) is the most effective way of using BERT in most of the GEC corpora that we used in the experiments.
We also show that the performance of GEC improves further by combining the BERT-fuse mask and BERT-fuse GED methods.
The best-performing model achieves state-of-the-art results on the BEA-2019 and CoNLL-2014 benchmarks.

\section{Related Work}

Studies have reported that a MLM can improve the performance of GEC when it is employed either as a re-ranker \cite{chollampatt2019csgec, kaneko-etal-2019-tmu} or as a filtering tool \cite{asano-etal-2019-aip, kiyono-etal-2019-empirical}.
EncDec-based GEC models combined with MLMs can also be used in combination with these pipeline methods.
\citet{kantor-etal-2019-learning} and \citet{awasthi-etal-2019-parallel} proposed sequence labeling models based on correction methods.
Our method can utilize the existing EncDec GEC knowledge, but these methods cannot be utilized due to the different architecture of the model.
Besides, to the best of our knowledge, no research has yet been conducted that incorporates information of MLMs for effectively training the EncDec GEC model.

MLMs are generally used in downstream tasks by fine-tuning \cite{liu2019fine, zhang2019pretraining}, however, \citet{Zhu2020IncorporatingBI} demonstrated that it is more effective to provide the output of the final layer of a MLM to the EncDec model as contextual embeddings.
Recently, \citet{weng2019acquiring} addressed the mismatch problem between contextual knowledge from pre-trained models and the target bilingual machine translation.
Here, we also claim that addressing the gap between grammatically correct raw corpora and GEC corpora can lead to the improvement of GEC systems. 

\section{Methods for Using Pre-trained MLM in GEC Model}
In this section, we describe our approaches for incorporating a pre-trained MLM into our GEC model.
Specifically, we chose the following approaches: 
(1) initializing a GEC model using BERT; 
(2) using BERT output as additional features for a GEC model, and 
(3) using the output of BERT fine-tuned with the GEC corpora as additional features for a GEC model.

\noindent
\subsection{BERT-init}
We create a GEC EncDec model initialized with BERT weights.
This approach is based on \citet{lample2019cross}.
Most recent state-of-the-art methods use pseudo-data, which is generated by injecting pseudo-errors to grammatically correct sentences.
However, note that this method cannot initialize a GEC model with pre-trained parameters learned from pseudo-data.

\noindent
\subsection{BERT-fuse}
We use the model proposed by \citet{Zhu2020IncorporatingBI} as a feature-based approach (BERT-fuse).
This model is based on Transformer EncDec architecture.
It takes an input sentence ${\boldsymbol{\rm X}} = (x_1, ... , x_n)$, where $n$ is its length.
$x_i$ is $i$-th token in ${\boldsymbol{\rm X}}$.
First, BERT encodes it and outputs a representation ${\boldsymbol{\rm B}} = (b_1, ... , b_n)$.
Next, the GEC model encodes ${\boldsymbol{\rm X}}$ and ${\boldsymbol{\rm B}}$ as inputs.
$h_i^l \in {\boldsymbol{\rm H}}$ is the $i$-th hidden representation of the $l$-th layer of the encoder in the GEC model.
$h^0$ stands for word embedding of an input sentence ${\boldsymbol{\rm X}}$.
Then we calculate $\tilde{h}_i^l$ as follows:
\begin{align}
 \tilde{h}_i^l = \frac{1}{2}(A_h(h_i^{l-1}, {\boldsymbol{\rm H}}^{l-1}) + A_b(h_i^{l-1}, {\boldsymbol{\rm B}}^{l-1})) \label{eq:enc_attn}
\end{align}
where $A_h$ and $A_b$ are attention models for the hidden layers of the GEC encoder ${\boldsymbol{\rm H}}$ and the BERT output ${\boldsymbol{\rm B}}$, respectively.
Then each $\tilde{h}_i^l$ is further processed by the feedforward network $F$ which outputs the $l$-th layer ${\boldsymbol{\rm H}}^{l} = (F(\tilde{h}_1^l), ... , F(\tilde{h}_n^l))$.
The decoder's hidden state $s_t^l \in \boldsymbol{\rm S}$ is calculated as follows:
\begin{align}
 \hat{s}_t^l &= A_s(s_t^{l-1}, {\boldsymbol{\rm S}}_{< t+1}^{l-1}) \\
 \tilde{s}_i^l &= \frac{1}{2}(A_h(\hat{s}_i^{l-1}, {\boldsymbol{\rm H}}^{l-1}) + A_b(\hat{s}_i^{l-1}, {\boldsymbol{\rm B}}^{l-1})) \label{eq:dec_attn} \\
 s_t^l &= F(\tilde{s}_t^l)
\end{align}
Here, $A_s$ represents the self-attention model.
Finally, $s_t^L$ is processed via a linear transformation and softmax function to predict the $t$-th word $\hat{y}_t$.
We also use the drop-net trick proposed by \citet{Zhu2020IncorporatingBI} to the output of BERT and the encoder of the GEC model.


\noindent
\subsection{BERT-fuse Mask and GED}
The advantage of the BERT-fuse is that it can preserve pre-trained information from raw corpora, however, it may not be adapted to either the GEC task or the task-specific distribution of inputs.
The reason is that in the GEC model, unlike the data used for training BERT, the input can be an erroneous sentence.
To fill the gap between corpora used to train GEC and BERT, we additionally train BERT on GEC corpora (BERT-fuse mask) or fine-tune BERT as a GED model (BERT-fuse GED) and use it for BERT-fuse.
GED is a sequence labeling task that detects grammatically incorrect words in input sentences~\cite{rei-yannakoudakis-2016-compositional,kaneko-etal-2017-grammatical}.
Since BERT is also effective in GED~\cite{bell-etal-2019-context,Kaneko2019MultiHeadMA}, it is considered to be suitable for fine-tuning to take into account grammatical errors.

\begin{table}[]
\centering
\scalebox{0.75}{
\begin{tabular}{lrr}
\toprule
\bf GEC model \\
\midrule
Model Architecture & Transformer (big) \\
Number of epochs & 30 \\
Max tokens & 4096 \\
Optimizer & Adam \\
 & ($\beta_1 = 0.9, \beta_2 = 0.98, \epsilon = 1 \times 10^{-8}$) \\
Learning rate & $3 \times 10^{-5}$ \\
Min learning rate & $1 \times 10^{-6}$ \\
Loss function & label smoothed cross-entropy \\
 & ($\epsilon_{ls}=0.1$) \\
 & \cite{44903} \\
Dropout & 0.3 \\
Gradient Clipping & 0.1 \\
Beam search & 5 \\
\midrule
\bf GED model\\
\midrule
Model Architecture & BERT-Base (cased) \\
Number of epochs & 3 \\
Batch size & 32 \\
Max sentence length & 128 \\
Optimizer & Adam \\
 & ($\beta_1 = 0.9, \beta_2 = 0.999, \epsilon = 1 \times 10^{-8}$) \\
Learning rate & $4{\rm e}-5$ \\
Dropout & 0.1 \\
\bottomrule
\end{tabular}
}%
\caption{Hyperparameters values of GEC model and Fine-tuned BERT.}
\label{tab:params}
\end{table}

\section{Experimental Setup}
\subsection{Train and Development Sets}
We use the BEA-2019 workshop\footnote{\url{https://www.cl.cam.ac.uk/research/nl/bea2019st/}}~\cite{bryant-etal-2019-bea} official shared task data as training and development sets.
Specifically, to train a GEC model, we use W\&I-train \cite{Granger-1998, yannakoudakis-etal-2018}, NUCLE \cite{dahlmeier-etal-2013-building}, FCE-train \cite{yannakoudakis-etal-2011-new} and Lang-8 \cite{mizumoto-etal-2011-mining} datasets.
We use W\&I-dev as a development set.
Note that we excluded sentence pairs that were not corrected from the training data. 
To train BERT for BERT-fuse mask and GED, we use W\&I-train, NUCLE, and FCE-train as training, and W\&I-dev was used as development data.

\subsection{Evaluating GEC Performance}
In GEC, it is important to evaluate the model with multiple datasets \cite{mita-etal-2019-cross}.
Therefore, we used GEC evaluation data such as W\&I-test, CoNLL-2014 \cite{ng-etal-2014-conll}, FCE-test and JFLEG \cite{napoles-sakaguchi-tetreault:2017:EACLshort}.
We used ERRANT evaluation metrics \cite{felice-etal-2016-automatic, bryant-etal-2017-automatic} for W\&I-test, ${\rm M}^2$ score \cite{dahlmeier-ng-2012-better} for CoNLL-2014 and FCE-test sets, and GLEU \cite{napoles-EtAl:2015:ACL-IJCNLP} for JFLEG.
All our results (except ensemble) are the average of four distinct trials using four different random seeds.

\begin{table*}[ht]
 \centering
 \scalebox{0.92}{
 \begin{tabular}{lrrrrrrrrrr}
 \toprule
 & \multicolumn{3}{c}{BEA-test (ERRANT)} & \multicolumn{3}{c}{CoNLL-14 (${\rm M^2}$)} & \multicolumn{3}{c}{FCE-test (${\rm M^2}$)} & {JFLEG} \\
 & \bf P & \bf R & $\bf F_{0.5}$ & \bf P & \bf R & $\bf F_{0.5}$ & \bf P & \bf R & $\bf F_{0.5}$ & \bf GLEU \\
 \midrule
 w/o BERT & 51.5 & 43.2 & 49.6 & 59.2 & 31.2 & 50.2 & 61.7 & 46.4 & 57.9 & 52.7 \\
 BERT-init & 55.1 & 43.7 & 52.4 & 61.3 & 31.5 & 51.4 & 62.4 & 46.9 & 58.5 & 53.0\\
 BERT-fuse & 57.5 & \bf 44.9 & 54.4 & 62.3 & 31.3 & 52.0 & 64.0 & 47.6 & 59.8 & 54.1 \\
 BERT-fuse mask & 57.1 & 44.7 & 54.1 & 62.9 & 32.2 & 52.8 & 64.3 & 48.1 & 60.2 & 54.2 \\
 BERT-fuse GED & \bf 58.1 & 44.8 & \bf 54.8 & \bf 63.6 & \bf 33.0 & \bf 53.6 & \bf 65.0 & \bf 49.6 & \bf 61.2 & \bf 54.4 \\
 \midrule
 w/o BERT & 66.1 & 59.9 & 64.8 & 68.5 & 44.8 & 61.9 & 56.5 & 48.1 & 54.9 & 61.0 \\
 BERT-fuse & 66.6 & 60.0 & 65.2 & 68.3 & \bf 45.7 & 62.1 & 59.7 & \bf 48.5 & \bf 57.0 & 61.2 \\
 BERT-fuse mask & 67.0 & 60.0 & 65.4 & 68.8 & 45.3 & 62.3 & 59.7 & 47.1 & 56.6 & 61.2 \\
 BERT-fuse GED & \bf 67.1 & \bf 60.1 & \bf 65.6 & \bf 69.2 & 45.6 & \bf 62.6 & \bf 59.8 & 46.9 & 56.7 & 61.3 \\
 \citet{lichtarge-etal-2019-corpora} & - & - & - & 65.5 & 37.1 & 56.8 & - & - & - & \bf 61.6 \\
 \citet{awasthi-etal-2019-parallel} & - & - & - & 66.1 & 43.0 & 59.7 & - & - & - & 60.3 \\
 \citet{kiyono-etal-2019-empirical} & 65.5 & 59.4 & 64.2 & 67.9 & 44.1 & 61.3 & - & - & - & 59.7 \\
 \midrule
 BERT-fuse GED + R2L & 72.3 & \bf 61.4 & 69.8 & \bf 72.6 & \bf 46.4 & \bf 65.2 & 62.8 & 48.8 & 59.4 & 62.0 \\
 \citet{lichtarge-etal-2019-corpora} & - & - & - & 66.7 & 43.9 & 60.4 & - & - & - & \bf 63.3 \\
 \citet{grundkiewicz-etal-2019-neural} & 72.3 & 60.1 & 69.5 & - & - & 64.2 & - & - & - & 61.2\\
 \citet{kiyono-etal-2019-empirical}$^{*}$ & \bf 74.7 & 56.7 & \bf 70.2 & 72.4 & 46.1 & 65.0 & - & - & - & 61.4 \\
 \bottomrule
 \end{tabular}}
 \caption{Results of our GEC models. The top group shows the results of the single models without using pseudo-data and/or ensemble. The second group shows the results of the single models using pseudo-data. The third group shows ensemble models using pseudo-data. {\bf Bold} indicates the highest score in each column. * reports the state-of-the-art scores for BEA test and CoNLL 2014 for two separate models: models with and without SED. We filled out a single line with the results from such two separate models.}
  \label{tab:gec_result}
\vspace{-1.5mm}
\end{table*}

\subsection{Models}
Hyperparameter values for the GEC model is listed in Table \ref{tab:params}.
For the BERT initialized GEC model, we provided experiments based on the open-source code\footnote{{\url{https://github.com/facebookresearch/XLM}}}.
For the BERT-fuse GEC model, we use the code provided by \citet{Zhu2020IncorporatingBI}\footnote{\url{https://github.com/bert-nmt/bert-nmt}}.
While the training the GEC model, the model was evaluated on the development set and saved every epoch.
If loss did not drop at the end of an epoch, the learning rate was multiplied by 0.7.
The training was stopped if the learning rate was less than the minimum learning rate or if the learning epoch reached the maximum epoch number of 30.

Training BERT for BERT-fuse mask and GED was based on the code from \citet{Wolf2019HuggingFacesTS}\footnote{\url{https://github.com/huggingface/transformers}}.
The additional training for the BERT-fuse mask was done in the \citet{devlin-etal-2019-bert}'s setting.
Hyperparameter values for the GED model is listed in Table \ref{tab:params}.
We used the BERT-Base cased model, for consistency across experiments\footnote{\url{https://github.com/google-research/bert}}.
The model was evaluated on the development set.

\subsection{Pseudo-data}
We also performed experiments utilizing BERT-fuse, BERT-fuse mask, and BERT-fuse GED outputs as additional features to the pre-trained on the pseudo-data GEC model.
The pre-trained model using pseudo-data was initialized with the \textsc{PretLarge+SSE} model used in the \newcite{kiyono-etal-2019-empirical}\footnote{\url{https://github.com/butsugiri/gec-pseudodata}} experiments.
This pseudo-data is generated by probabilistically injecting character errors into the output~\cite{lichtarge-etal-2019-corpora} of a backtranslation~\cite{xie-etal-2018-noising} model that generates grammatically incorrect sentences from grammatically correct sentences~\cite{kiyono-etal-2019-empirical}.

\subsection{Right-to-left (R2L) Re-ranking for Ensemble}
We describe the R2L re-ranking technique incorporated in our experiments proposed by \citet{sennrich-etal-2016-edinburgh}, which proved to be efficient for the GEC task \cite{grundkiewicz-etal-2019-neural, kiyono-etal-2019-empirical}.
Standard left-to-right (L2R) models generate the $n$-best hypotheses using scores with the normal ensemble and R2L models re-score them.
Then, we re-rank the $n$-best candidates based on the sum of the L2R and R2L scores.
We use the generation probability as a re-ranking score and ensemble four L2R models and four R2L models.

\section{Results}

Table \ref{tab:gec_result} shows the experimental results of the GEC models.
A model trained on Transformer without using BERT is denoted as ``w/o BERT.''
In the top groups of results, it can be seen that using BERT consistently improves the accuracy of our GEC model.
Also, BERT-fuse, BERT-fuse mask, and BERT-fuse GED outperformed the BERT-init model in almost all cases.
Furthermore, we can see that using BERT considering GEC corpora as BERT-fuse leads to better correction results.
And the BERT-fuse GED always gives better results than the BERT-fuse mask.
This may be because the BERT-fuse GED is able to explicitly consider grammatical errors.
In the second row, the correction results are improved by using BERT as well.
Also in this setting, BERT-fuse GED outperformed other models in all cases except for the FCE-test set, thus, achieving state-of-the-art results with a single model on the BEA2019 and CoNLL14 datasets.
In the last row, the ensemble model yielded high scores on all corpora, improving state-of-the-art results by 0.2 points in CoNLL14.

\section{Analysis}
\subsection{Hidden Representation Visualization}
We investigate the characteristics of the hidden representations of vanilla (i.e., without any fine-tuning) BERT and BERT fine-tuned with GED.
We visualize the hidden representations of the same words from the last layer of BERT ${\boldsymbol{\rm H}}^{L}$. They were chosen depending on correctness in a different context, using the above models.
These target eight words\footnote{1. the 2. , 3. in 4. to 5. of 6. a 7. for 8. is} that have been mistaken more than 50 times, were chosen from W\&I-dev.
We sampled the same number of correctly used cases for the same word from the corrected side of W\&I-dev.

\begin{figure}[t]
  \begin{subfigure}[b]{0.237\textwidth}
    \includegraphics[width=\linewidth]{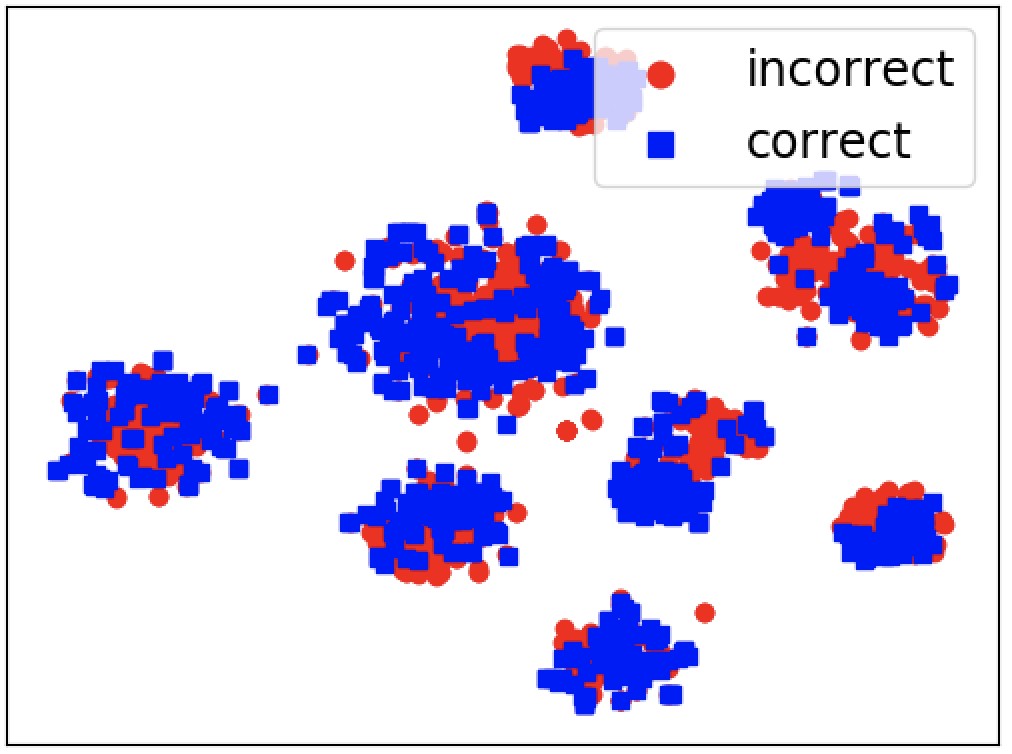}
    \caption{BERT}
    \label{fig:bert_fuse}
  \end{subfigure}
  \hfill
  \begin{subfigure}[b]{0.237\textwidth}
    \includegraphics[width=\linewidth]{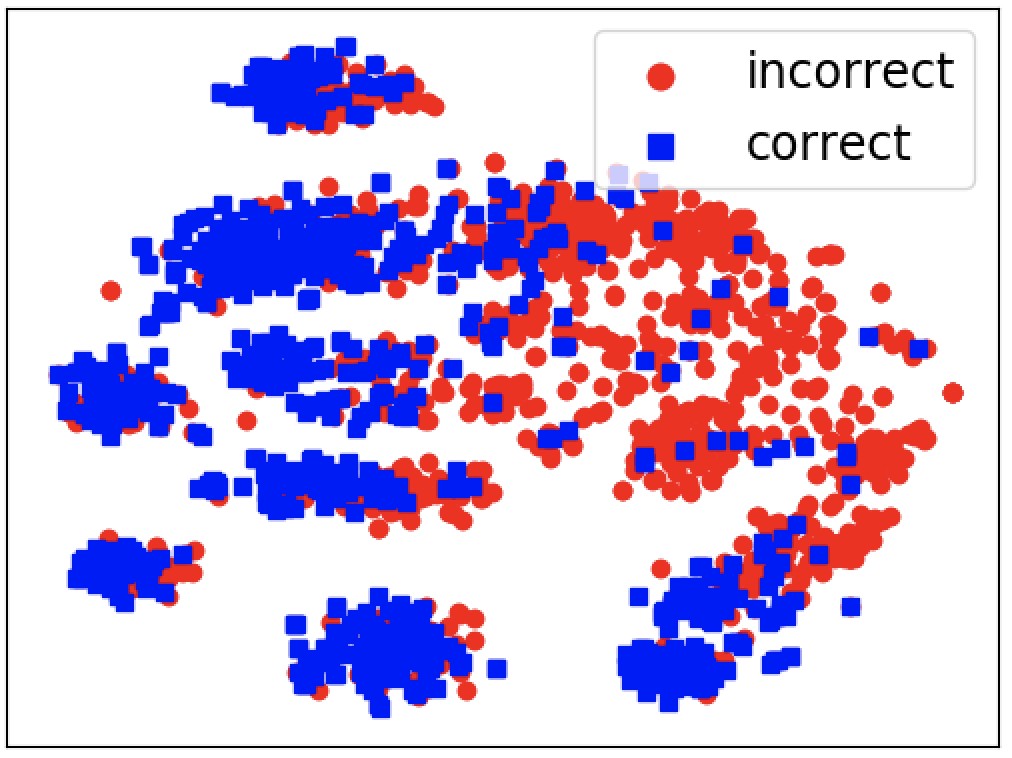}
    \caption{Fine-tuned BERT}
    \label{fig:ft_bert_fuse}
  \end{subfigure}\hfill
  \caption{Hidden representation visualization for encoded grammatically correct and incorrect words. }\label{fig:vizualization}
\vspace{-1.5mm}
\end{figure}

Figure \ref{fig:vizualization} visualizes hidden representations of BERT and fine-tuned BERT.
It can be seen that the vanilla BERT does not distinguish between correct and incorrect clusters.
The plotted eight words are gathered together, and it can be seen that hidden representations of the same word gather in the same place regardless of correctness.
On the other hand, fine-tuned BERT produces a vector space that demonstrates correct and incorrect words on different sides, showing that hidden representations take grammatical errors into account when fine-tuned on GEC corpora.
Moreover, it can be seen that the correct cases divided into 8 clusters, implying that BERT's information is also retained.

\subsection{Performance for Each Error Type}

We investigate the correction results for each error type.
We use ERRANT~\cite{felice-etal-2016-automatic, bryant-etal-2017-automatic} to measure $\rm F_{0.5}$ of the model for each error type.
ERRANT can automatically assign error types from source and target sentences.
We use single BERT-fuse GED and w/o BERT models without using pseudo-data for this investigation.

Table \ref{tab:err_type} shows the results of single BERT-fuse GED and w/o BERT models without using pseudo-data on most error types including all the top-5 frequent error types in W\&I-dev.
We see that BERT-fuse GED is better for all error types compared to w/o BERT.
We can say that the use of BERT fine-tuned by GED for the EncDec model improves the performance independently of the error type.

\begin{table}[t]
\centering
\scalebox{0.87}{
\begin{tabular}{lrr}
\toprule 
Error type & BERT-fuse GED & w/o BERT \\
\midrule
PUNCT   & \bf 40.2 & 36.8 \\
OTHER   & \bf 20.4 & 19.1 \\
DET    & \bf 48.8 & 45.4 \\
PREP    & \bf 36.7 & 34.8 \\
VERB:TENSE & \bf 36.0 & 34.1 \\
\bottomrule
\end{tabular}}
\caption{The result of single Fine-tuned BERT-fuse and w/o BERT models without using pseudo-data on most error types including all the top-5 frequent types of error in W\&I-dev}
\label{tab:err_type}
\end{table}

\section{Conclusion}

In this paper, we investigated how to effectively use MLMs for training GEC models.
Our results show that BERT-fuse GED was one of the most effective techniques when it was fine-tuned with GEC corpora.
In future work, we will investigate whether BERT-init can be used effectively by using methods to deal with catastrophic forgetting.

\section*{Acknowledgments}
This work was supported by JSPS KAKENHI Grant Number 19J14084 and 19H04162.
We thank everyone in Inui and Suzuki Lab at the Tohoku University and Language Information Access Technology Team of RIKEN AIP.
We thank the anonymous reviewers for their valuable comments.

\bibliography{acl2020}
\bibliographystyle{acl_natbib}



\end{document}